# Application of Deep Learning Long Short-Term Memory in Energy Demand Forecasting

Nameer Al Khafaf[✉][0000-0002-4295-6946], Mahdi Jalili and Peter Sokolowski

Electrical and Biomedical Engineering, RMIT University, Melbourne, Australia

nkhafaf@gmail.com, mahdi.jalili@rmit.edu.au,
peter.sokolowski@rmit.edu.au

**Abstract.** The smart metering infrastructure has changed how electricity is measured in both residential and industrial application. The large amount of data collected by smart meter per day provides a huge potential for analytics to support the operation of a smart grid, an example of which is energy demand forecasting. Short term energy forecasting can be used by utilities to assess if any forecasted peak energy demand would have an adverse effect on the power system transmission and distribution infrastructure. It can also help in load scheduling and demand side management. Many techniques have been proposed to forecast time series including Support Vector Machine, Artificial Neural Network and Deep Learning. In this work we use Long Short Term Memory architecture to forecast 3-day ahead energy demand across each month in the year. The results show that 3-day ahead demand can be accurately forecasted with a Mean Absolute Percentage Error of 3.15%. In addition to that, the paper proposes way to quantify the time as a feature to be used in the training phase which is shown to affect the network performance.

**Keywords:** Deep Learning, Long Short-Term Memory, Demand Forecasting,

## 1      Introduction

The introduction of smart meters technology in recent years have shaped the metering infrastructure industry providing a smart, efficient and regular monitoring of energy consumption. Smart meters have been deployed in almost all residential and industrial applications around the world and in Australia giving the potential for metering intelligence and analytics [1, 2]. The smart meter usually captures aggregate of energy consumption in either a 15 or 30 minutes window creating a historic profile of energy consumption for each user.

Energy consumption is mainly driven by the energy consumer actions and behaviours which are affected by the consumer preferences. Since consumers' preferences are likely to change over time, this introduces uncertainties in the daily energy consumption pattern. Other factors influencing energy consumption include economic situations, climate change, holidays, working days, time periods and social and behav-ioural aspect[3]. In addition to that electricity demand is on the rise with increased population and introduction of many appliances such as dish washer and cloth dryer into the household. Another contributor to the increase in energy consumption is climate change which contributes to a spike in energy consumption for approximately 40 hours a year or 0.5% according to a report by Powercor Australia [4]. The supply side may or may not be able to handle the spike in demand if it has not been properly planned. One way to manage the spike in demand is to increase electrical assets such as generation, distribution and transmission infrastructure to accommodate the increase in energy consumption. However this increase in asset is not suitable as the cost of installing the electrical infrastructure far outweigh the benefits of meeting the spike in demand as it only happens about 0.5% of the year. Another way of managing the sudden increase in energy consumption is through energy management strategies applied at the demand side [5, 6]. In order to properly and effectively manage peak energy consumption in the short term, an accurate load forecasting is required.

Load forecasting is one of the most important analytics for the smart grid as it provides a prediction of what would be the likely energy demand in the future with a margin of error allowing for a timely decision making[7]. The purpose of demand forecasting depends on the prediction period such as short, medium or long term. Short, medium and long term forecasting are defined as being less than 1 week, between 1 week to one year and more than one year respectively[8]. Short term forecasting can be used as an input into demand side management framework to help better addressing high peak consumption due to specific events such as heat wave or blizzard depending on the season. Medium term forecasting is mostly used for load scheduling and maximizing power distribution and transmission asset utilization. Whereas long term forecasting focuses on identifying time period where the demand will be the lowest to plan for maintenance or shut down for upgrade. Long term is also used to plan for upgrading the power network due to a constant and permanent increase in demand that is not prone to seasonal or daily fluctuations[9].

There are many applications for load forecasting based on statistical and machine learning technologies that have been developed in the literature[10]; time series and artificial neural networks are most common techniques for short term and medium term forecasting[11-13]. Other short term forecasting are based on deep learning Long Short-Term Memory (LSTM) approach which has been proven in [14, 15] to be effective compared to traditional approach. LSTM has also been widely used and proven to be effective in short and medium term forecasting [16, 17].

A recent study by [18] to develop a high precision ANN for load forecasting (DeepEnergy) conclude that the proposed technique exceeds the traditional LSTM network in terms of the Mean Absolute Percentage Error (MAPE) for a 3-day ahead forecast. However in their implementation of LSTM, they did not consider other features to effectively train the LSTM network. Moreover they used data from two different months for training and data from a third month for forecasting. This may affect the prediction accuracy of LSTM as consumption can be different from one month to another. In another study conducted by [2] using dynamic neural network to load forecast which seems to be giving a good accuracy, however similar to the work by [18], the experiment does not consider other features to improve the accuracy of the model rather depend solely on the historic energy consumption.

In this paper we propose a LSTM deep learning to forecast energy demand for clusters of energy users in the short and medium term. The difference between our work and the work in the literature is in the way we pre-process the raw data for training and the use of two more features in addition to the historic energy consumption to improve the forecasting process. The first trial is to forecast 3-day ahead energy consumption in each month and the second trial is to forecast 15 days ahead energy consumption in a year. This work is anticipated to be an important step toward developing a LSTM model to accurately forecast peak energy consumption in the short and medium term. The model can be used by utilities to prepare for spike in energy and hence power demand during a heat wave.

## 2    Research Approach

The approach aims at forecasting a 3-day ahead energy consumption of clusters of energy users and accurately predicts peak consumption occurring during the 3 days prediction window. This will allow utilities to take actions to manage the spike in demand if the forecasted demand is higher than the capacity of the supply side. In addition to the 3-day ahead prediction, we attempt to forecast 15 days ahead energy consumption by training the LSTM) on larger data. The LSTM we used has the architecture defined in [3] which consists of 3 gate units namely, input, output and forget gate to form one memory cell. Gates control the flow of the energy consumption time series inside the LSTM unit whereas the cell record dependencies between values of the time series. Each gate uses a sigmoid activation function $\Phi_S$ whereas the cell state uses hyperbolic tangent activation function $\Phi_T$.

Given an input time series $y_t$ the LSTM has to learn the input weights $I$, the recurrent weights $R$ and the bias $b$ where $I$, $R$ and $b$ are 4-by-1 column vectors with elements correspond to input, output and forget gate and the cell state. The input $i$, forget $f$ and output $o$ gates at time step $t$ are written as:

$$i_t = \Phi_S(I_i x_t + R_i h_{t-1} + b_i) \quad (1)$$

$$f_t = \Phi_S(I_f x_t + R_f h_{t-1} + b_f) \quad (2)$$

$$o_t = \Phi_S(I_o x_t + R_o h_{t-1} + b_o) \quad (3)$$

Where $h_{t-1}$ is the output of the LSTM cell at the previous time step. Similarly, the cell state $c$ is written as:

$$c_t = \Phi_T(I_c x_t + R_c h_{t-1} + b_c) \quad (4)$$

At each time step, the LSTM uses the time series to compute $c_t$ and $h_t$ which are then fed into a regression layer to predict the time series value. We used the root mean square error as a metric performance during training.

## 3 Data Processing

### 3.1 The Data

The energy consumption of 609 anonymous households in Victoria Australia have been collected by Power Distribution Company using smart meters every 30 minutes for a full year starting from 9 March 2015 and ending on 8 March 2016. The raw dataset came in 9 files and each file comprised of the energy consumption of all energy consumers in each period of time. Table 1 presents information on the raw dataset used for the research.

**Table 1.** Smart meter energy consumption raw data

| Data File No. | Start Date | End Date | Total No. of Sample Points |
|---|---|---|---|
| 9 | 9-Mar-15 | 19-Apr-15 | 1,227,744 |
| 1 | 20-Apr-15 | 31-May-15 | 1,227,744 |
| 2 | 1-Jun-15 | 11-Jul-15 | 1,198,512 |
| 3 | 12-Jul-15 | 22-Aug-15 | 1,227,744 |
| 4 | 23-Aug-15 | 2-Oct-15 | 1,198,512 |
| 5 | 3-Oct-15 | 14-Nov-15 | 1,256,976 |
| 6 | 15-Nov-15 | 27-Dec-15 | 1,256,976 |
| 8 | 28-Dec-15 | 6-Feb-16 | 1,198,512 |
| 7 | 7-Feb-16 | 8-Mar-16 | 906,192 |

Combining all data files into one produces a 3-by-10,698,912 matrix where the first, second and third column corresponds to consumer label, consumption time and energy consumption respectively. The next step is to pre-process the raw data into a feature vector for clustering and then forecasting.

### 3.2 Data Pre-processing and Analysis

We then reorganize the data into an energy consumption matrix $A$ 17520-by-609 where columns are energy consumptions in ascending order and rows are the energy consumers identification from the dataset. In addition, we create a time column vector $B$ 17520-by-1 to preserve the time instances corresponding to energy consumption.

Few time instances were missing for some of the energy consumers due to either collecting the data days earlier compared to other consumers or after. These time instances were removed from the energy consumption matrix to avoid dealing with empty cells when training the neural network hence reducing the size of Matrix $A$ to 17496-by-609.

### 3.3 Clustering

Forecasting the load of individual energy consumer is often impracticable in residential sector as each consumer contributes a small proportion to the loading of the transformer or the connection point unless the consumer is a large industrial load which can be treated separately. As a result we employ a technique to cluster the dataset into optimal number of clustering thus reducing the number of loads into manageable time series for forecasting. The technique uses self-organizing maps to cluster the dataset from 2 to n clusters and use mathematical equation at each iteration to decide whether the number of clusters is optimal or not. We found that the optimal number of clusters for this dataset is 4 clusters which are plotted in fig. 1. The clustering technique is discussed in another paper and is outside the scope of this work.

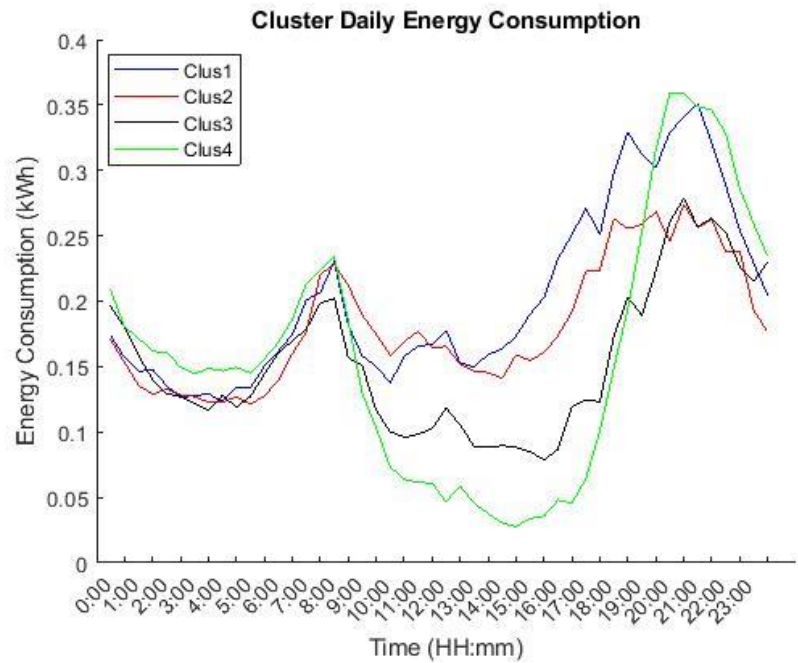

**Fig. 1.** Representative cluster consumption profile of 609 energy consumers within the dataset

### 3.4 Features Definition

We have identified three features that affect the forecast error and improve predications namely; historic energy consumption, daily temperature and time of day. Historic energy consumption provides insight into how much energy is being used in a day in each month at specific time of the days. This allows the forecast model to learn from historic consumption to predict future ones.

The daily temperature affects the consumer behavior to use less or more energy depending on how hot or cold the weather is at a specific time instance. Temperatures in the range of 19 to 25 degrees C are comfortable temperature that would unlikely to influence the energy consumption; however any temperature higher or lower than the comfortable temperature range would likely to influence the energy consumption. Fig. 2a shows a typical plot of a temperature profile across the day and fig. 3 shows how the clusters energy consumption change during a hot day when the temperature is higher than the comfortable temperature compared to a normal day. This is evident from the energy consumption range of 0.1 to 1.2 kWh in fig. 3a during a hot day compared to the energy consumption range of 0.1 to 0.35 kWh during a normal day in fig. 3c

Time affects energy consumer as it represent the instance when a specific energy is consumed. This is usually different for each consumer as the consumption will b affect by daily behavior such as scheduling of regular loads such laundry, dish washing, TV and other electronics devices. Another factor that impact consumption on different is whether the house become non-occupant during a certain time of the day where parent are at work and children at school. This can change from one family to another. The day of the week is also important as the consumer behavior is likely to be different on weekdays compared to weekends. As a result it is vital to propose a time feature where each time instances in a 7 day is unique and each day in a week is unique as week. We construct the feature vector from the time instances where each day is given a number from 1 to 7, where 2-6 represent weekdays and 1 and 7 represent weekends, and each 30-minute interval in the day is given a number from 1 to 48 where corresponds to 00:00 and 48 corresponds to 23:00. The time instance then becomes a factor of day and time by appending the time number to the end of the day number. For example, a 8:00 am on Tuesday is written as 318 where 3 corresponds to Tuesday and 18 corresponds to 8:00am. We then divide this number by the highest value which is 748 corresponding to Saturday at 23:00 or 11:30pm to get a time feature value between 0 and 1. Fig. 2b is a plot of the time feature profile across the week.

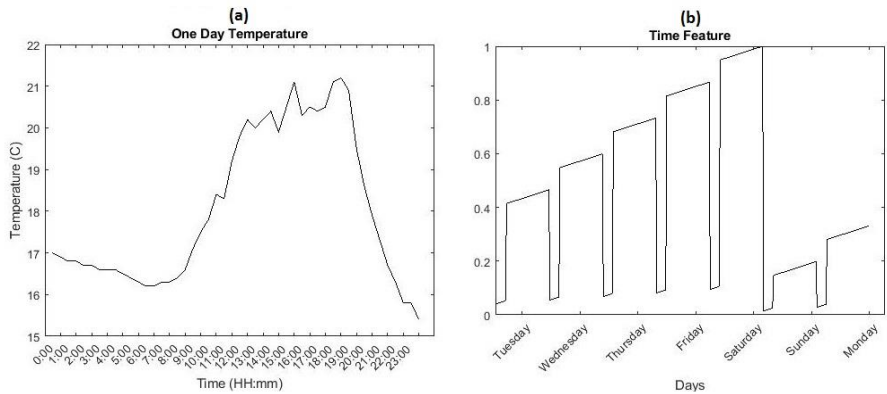

**Fig. 2.** (a) Temperature profile in degree C across the day for 0 March 2015; (b) Time feature profile to across the week giving each 30-minute time instance a unique number

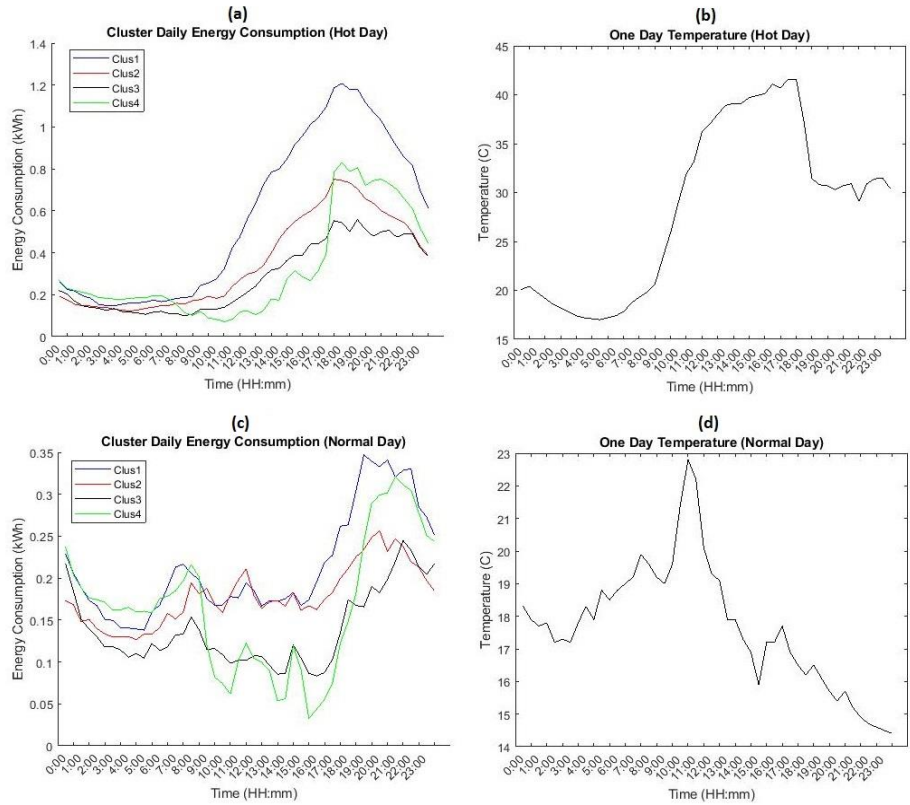

**Fig. 3.** (a) Clusters energy consumption on a hot day; (b) the temperature across the day for 13 January 2016; (c) clusters energy consumption on a normal day; (d) the temperature across the day for 16 October 2015 unique number

## 4 Experimental Results

We conducted the simulation over three experiments where each experiment tests different hypothesis. The first experiment tests the effectiveness of LSTM short-term forecasting capabilities by comparing performance error of 3-day ahead forecasts for cluster 1 across different months of the year using the historic energy consumption along with either the temperature or the time feature and by using all three features together. The second experiment uses both time and temperature features to forecast energy consumption for clusters 2,3 and 4 and third experiment aims to forecast 15 days ahead in a year. In all cases the dataset is divided into a ratio of 70,10 and 10 for training, validation and testing respectively. In the case of 3-day ahead forecasts the dataset corresponds to daily consumptions in each month, however in the case of 15 days ahead forecast the dataset corresponds to the daily consumption for the entire year.

## 4.1 Experiment 1 – 3-day ahead forecast ( 2/3 Features)

We use both the MAPE and the RMSE in testing the performance of the LSTM which is depicted in fig. 4. It can be seen from the figure that the lowest error in terms of MAPE is achieved by the 3 features forecast in October (fig. 4a) and the lowest error in terms of RMSE is achieved by the 2 features (temperature feature) forecast in April (fig. 4b). On average the 3 features forecast across all months is about 5% in terms of MAPE and 5.2% in terms of RMSE. While the average RMSE for 2 features forecast scored slightly lower (less than 0.5%) than the 3 features however the average MAPE for the 3 features is considerably lower (2%) than 2 features

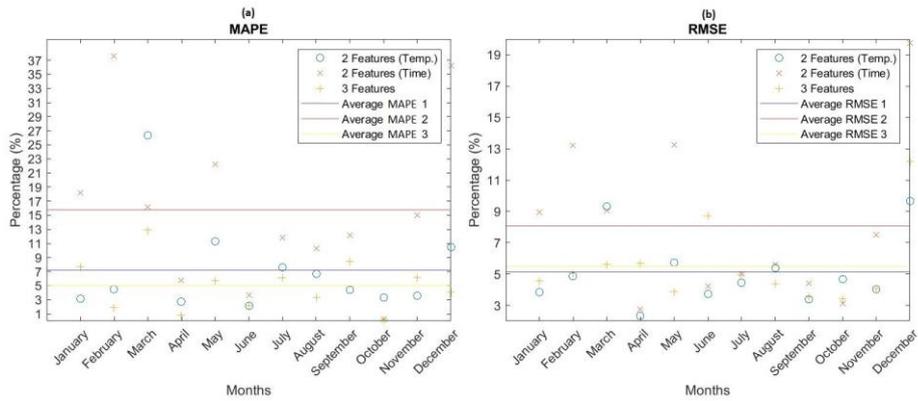

**Fig. 4.** The mean absolute percentage error fig.(a) and the root mean square error fig.(b) for 3-day ahead prediction for LSTM in each month and the average MAPE across all months for 2 and 3 features

Figure 5 shows the 3-day ahead energy consumption forecasts for April, June, March and November. As noted from figure 4 and 5, using 3 features gives a better forecast performance compared to using 2 features only.

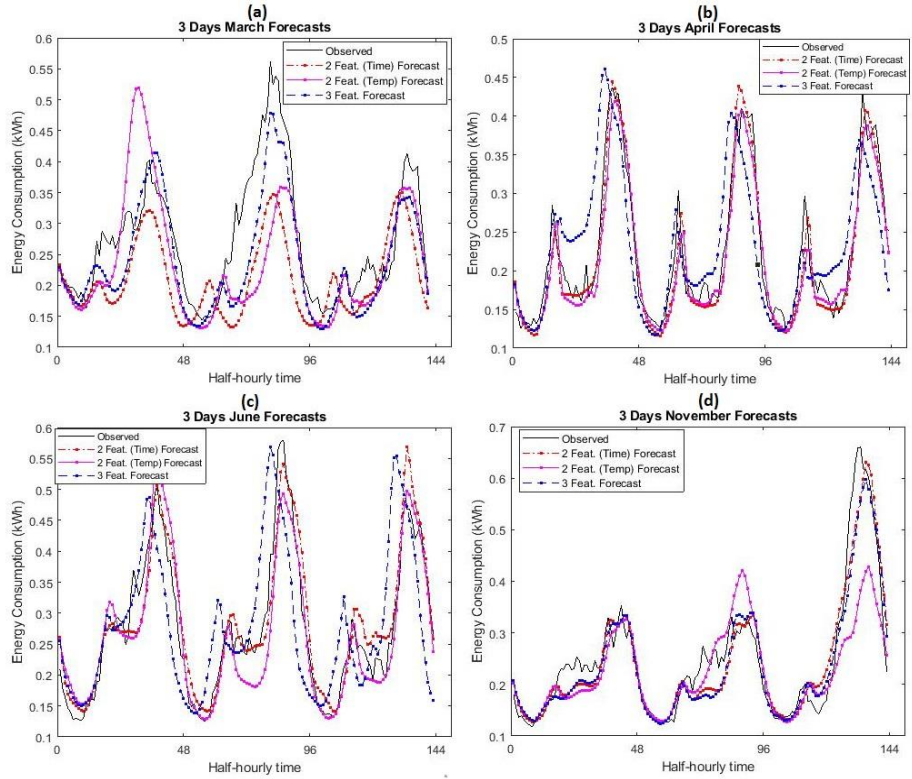

**Fig. 5.** 3-day ahead forecasts for 2 and 3 features in the months of April fig. (a), June fig.(b), March fig. (c) and November fig.(d)

### 4.2    Experiment 2 – 3-day ahead forecast (3 Features on 3 remaining clusters)

In this experiment we use the monthly energy consumption from the dataset to train the LSTM to forecast 3-day ahead for the remaining three clusters and using the 3 features defined in previous section. Fig. 6 shows the forecast energy consumption compared with the actual. It can be noted from this figure that the peak energy consumption can be accurately predicted in September for different energy consumption profile. As a result, the proposed methodology, features and LSTM architecture can be used to accurately forecast 3-day ahead energy consumption of various pattern or shape

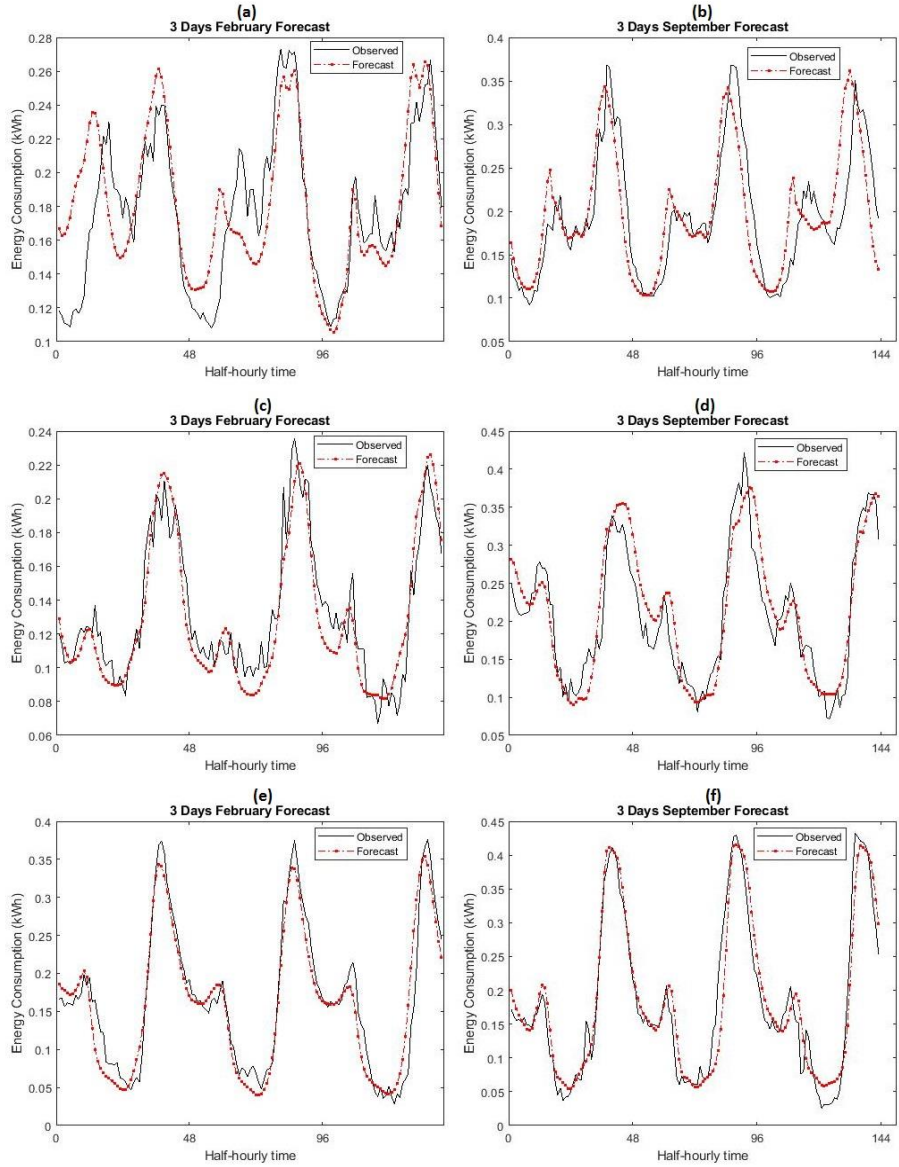

**Fig. 6.** 3-day ahead forecasts using 3 features in the months of February and September for cluster 1 fig.(a) and (b), cluster 3 fig.(c) and (d) and cluster 4 fig,(e) and (f)

### 4.3   Experiment 3 – 15-day ahead forecast (3 Features)

In this experiment we use the energy consumption for the whole year to train the LSTM to forecast 15-day ahead energy demand. Fig. 7 depicts the forecast energy consumption compared with the actual consumption for cluster 4. It can be noted from the figure that the energy consumption is forecasted with acceptable forecasting error of 3.7624% and 3.61% in terms of MAPE and RMSE respectively. A 15 days ahead prediction can very useful for utilities to determine whether peak consumption is likely to occur during the 15 days so they can plan for it accordingly.

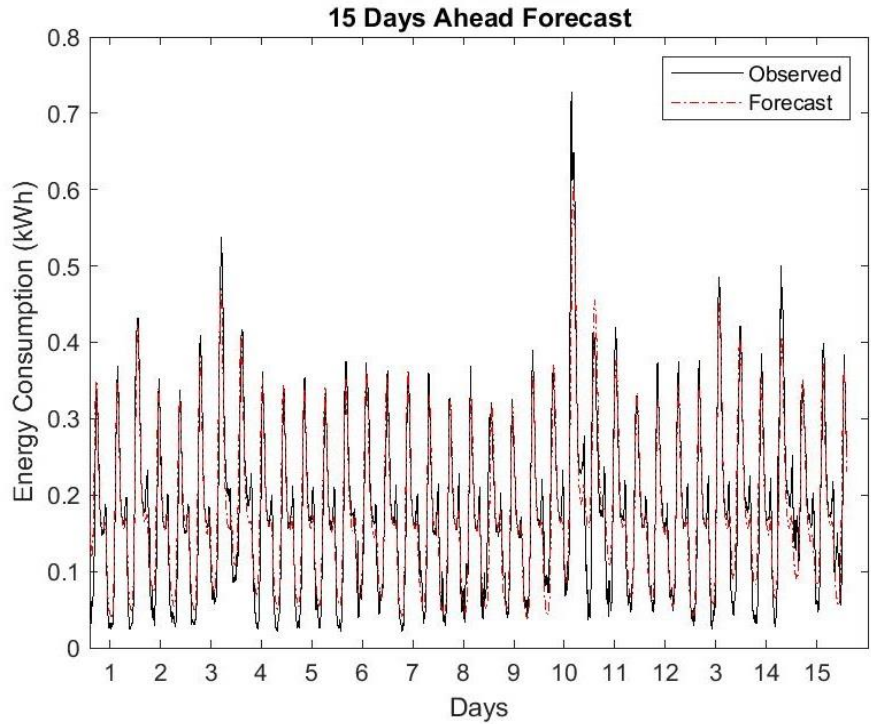

**Fig. 7.** 15 days ahead energy consumption forecast for cluster 4 with MAPE of 3.7624% and RMSE of 3.61%

## 5    Conclusions and Future Work

This work focuses at the potential of using deep learning LSTM to forecast 3 and 15 days ahead energy demand of different load profiles. The outcome of the research suggests that LSTM is a strong architecture for both short and medium term forecasting. We have also shown that defining effective features improve the forecasting model. As load and energy demand is critical for demand side management, this work can provide utilities with the needed information to make decision on how to better manage peak energy demand with an error of 3.15%. As future works, we plan to improve the LSTM forecasting in short term and accurately forecast 3-month ahead by tweaking the existing architecture as well as develop LSTM deep learning network for long term forecasting.